# Direct GPU Energy Measurements of Batch-Level Sharing in Mixture-of-Experts and Recurrent Models

**Rafael Pina**[1] **and Varuna De Silva**[2*]
[1]Institute for Digital Technologies, Loughborough University London, London, United Kingdom
[2] Center for Neuromorphic Intelligence, Aston University, Birmingham, United Kingdom

## Abstract

Energy claims for efficient neural networks are often inferred from operation counts. We report direct GPU energy measurements on two workloads that batch very differently, recurrent policies in GPU-vectorised multi-agent reinforcement learning (MARL) and mixture-of-experts (MoE) language models, and ask one question of both: can the rows of a batch be made to share computation, across the batch and over time, at a measured saving? On recurrent policies the answer so far is no. Shared activity patterns exist but rarely last one step, masking a frozen policy fails, and trained-in masks succeed only because a fixed mask is equivalent to a narrow GRU, at 11× the energy of a small dense policy in our implementation. Test-time steering adds nothing beyond re-picking on either workload. On MoE models the answer is yes, and granularity decides how much: on Qwen3-30B-A3B, batch-aware routing reads 25–31% fewer expert weights within KL 0.05, with benchmark changes within noise at 500 items, and uses 12–18% less energy per token on a fused forward. We also introduce persistence-aware OEA, a one-parameter rule that keeps the batch-aware cut with about 40% fewer new experts per step than native routing at batch 32–64, though its energy has not yet been measured. These are pilot measurements on one GPU type, and we state their limits.

## 1 Introduction

A GPU saves energy when many rows of a batch do the same work: one instruction stream, one set of weights, a full tile of data. Methods that save operations per sample often break this condition. Under batching, per-sample sparsity in a language model's MLP layers vanishes [1]. MoE routing fragments microbatches and reduces weight reuse during decoding [2], and a token's cost depends on how the request is scheduled and batched [3, 4]. Operation counts therefore say little about joules [5, 6]. What matters is whether computation can be *shared*, across the rows of a batch and across successive steps, without harming the model.

This paper reports a pilot that tested that question directly, measuring GPU energy wherever it could. We began with recurrent policies in GPU-vectorised MARL, where the environment, policy and learner share one device [7], then moved to MoE language models, where routing is an explicit discrete decision. Several of our most useful findings are negative, and we report them in full. Our contributions are:

- **An energy map of vectorised MARL** (Section 4.1): the update takes 82% of energy at 4,096 environments, and Amdahl's law caps any cheaper policy network at 3.35×.
- **Controls that change how trained-in masks should be read** (Section 4.2): one fixed mask matches state-gated policies, a fixed mask is equivalent to a narrow GRU, and in our implementation masked networks are the most expensive tested.
- **Evidence against test-time steering** for this purpose (Section 4.3), on both frozen recurrent policies and MoE routing.

*Corresponding author. Emails: r.m.pina@lboro.ac.uk (R.P.); v.desilva@aston.ac.uk (V.D.S.).

- **Measured joules for batch-aware routing** (Section 4.4–4.5). In a targeted search we found no earlier energy measurement of these rules, which report latency and expert counts [8, 9].
- **Persistence-aware OEA** (Sections 3.3 and 4.6), showing that the batch and time axes need not conflict at the level of expert sets.

## 2 Background and related work

**MoE models and their cost at inference.** An MoE layer holds many parallel feed-forward experts, and a router sends each token to its top-$k$ [10, 11]. Models such as Mixtral, OLMoE and Qwen3 use this to grow capacity at fixed per-token compute [12, 13, 14]. A batch, however, must read the union of every token's experts. Adhinarayanan et al. show that this reuse fragmentation can push the feed-forward stage into a bandwidth-bound regime [2], and Lee et al. report that chunked prefill adds up to 39% memory traffic through redundant expert weight loads [4]. Sparse kernels such as MegaBlocks and ScatterMoE make the expert computation itself efficient [15, 16].

**The batch axis.** Lynx, OEA and XShare re-route tokens at run time so that a batch uses fewer experts, without retraining [8, 9, 17]. OEA reports 39% lower MoE-layer decode latency on Qwen3-30B at batch 16, and XShare up to 30% fewer expert activations. These rules look at one decode step at a time and report latency and activations, not energy.

**The time axis.** ReMoE fine-tunes the router towards recently used experts, raising reuse by 26% and cutting time per output token under offloading [18]. Sticky Routing trains with a consistency loss between adjacent tokens [19]. Temporally Extended MoE adds an option-critic controller with a deliberation cost [20, 21, 22], and Liang et al. find that routing consistency trades off against local load balance across 20 models [23]. These methods act per sequence and mainly avoid weight transfers when experts do not fit in GPU memory. In our experiments all weights stay in GPU memory, so those gains do not transfer directly. Table 1 summarises the split. In a targeted, non-systematic search we found no rule that acts on both axes at run time with measured energy.

Table 1: How the closest work relates to this paper, as reported by its authors.

| Work | Mechanism | Batch | Time | Reports |
|---|---|---|---|---|
| Lynx [17], OEA [8], XShare [9] | Run-time re-routing within a batch | yes | no | Latency, activations |
| ReMoE [18] | Router fine-tuning for reuse | no | yes | Reuse, latency |
| Sticky Routing [19] | Consistency loss in training | no | yes | Switches, perplexity |
| Temporally Extended MoE [22] | Option-critic layer controller | no | yes | Switch rate, accuracy |
| Polar Sparsity [1] | Selective kernels; shows MLP sparsity vanishes under batching | yes | no | Throughput |
| **This paper** | **OEA with a previous-step bonus; energy of re-routing** | **yes** | **yes** | **Joules, new experts, KL** |

**Recurrent policies.** Efficiency methods for agents, such as activity-sparse recurrent units [24], act per unit and per sample, which is hard to exploit on a batched GPU. We therefore ask whether a batch of agents shares activity patterns, whether they persist, and whether they can be trained in. We also tested steering, since vector interventions are a cheap way to change a frozen model, although simple baselines often outperform them for precise control [25].

## 3 Methodology

### 3.1 Workloads and measurement

Table 2 lists the workloads. Energy is read from the NVML cumulative energy counter at phase boundaries, with an idle baseline taken in the same session, and we report raw energy as the primary value because measured idle power varies between sessions. The routing comparisons use locked clocks and interleaved rounds, with native routing measured before and after each round so that a rule is compared with its immediate neighbours. Full protocols are in Appendix A.

Table 2: Workloads. All runs used NVIDIA RTX A6000 GPUs (300 W limit). Qwen3 is split across two cards.

| Workload | Model | Task and data | Measures |
|---|---|---|---|
| MARL energy | IPPO [26], GRU 128 | smacv2_10_units [27], 128–4,096 envs, 3 seeds | J per env-step, kernel time |
| MARL structure | GRU width 2,048, 16 tiles of 128 | SMAX [7, 28] 3m and 5m_vs_6m, 3–5 seeds | Signatures, wins, J per agent-step |
| MoE, coarse | OLMoE-1B-7B: 16 layers, 64 experts, top-8 | WikiText [29], 3 shards, $B$ = 16, 64 | $U$, KL, persistence |
| MoE, fine | Qwen3-30B-A3B: 48 layers, 128 experts, top-8 | WikiText, 3 shards, $B$ = 16, 32, 64 | $U$, KL, persistence, J per token |

### 3.2 Metrics

For MoE routing, let $S_t$ be the set of experts that a layer reads for the batch at decode step $t$. We report the mean $U = |S_t|$ against its native value $U_0$, the Jaccard persistence and the number of new experts per step, averaged over steps and layers:

$$J_t = \frac{|S_t \cap S_{t-1}|}{|S_t \cup S_{t-1}|}, \quad N_t = |S_t \setminus S_{t-1}|. \tag{1}$$

Quality is the mean KL divergence of the modified model's next-token distribution from the unmodified model's, per continuation token, with a budget of 0.05 nats, together with top-1 agreement, perplexity and benchmark accuracy on 500 items each of ARC-Easy and HellaSwag [30, 31]. Decoding is teacher-forced. Settings are chosen on shard 0 and repeated on all three shards. If a fraction $p$ of the work becomes $s$ times cheaper, the saving is bounded by Amdahl's law [32]:

$$S = \frac{1}{(1-p) + p/s} \le \frac{1}{1-p}. \tag{2}$$

For MARL, a row's **signature** is its $k$ most active tiles by L2 norm. **Coverage@64** is the share of rows whose signature is among the batch's 64 most common, and **persistence** is the share of rows whose signature is unchanged at the next step, excluding episode boundaries.

### 3.3 Routing rules, and persistence-aware OEA

We compare three published or simple rules. **R-margin** swaps a token's expert outside a target set for its best unused expert inside when the logit gap is below a threshold. **OEA** [8] keeps each token's top-$k_0$, takes the union over the batch, and fills the remaining slots only from that union. **XShare** [9] adds experts by batch-summed gate score up to a budget. OEA has no memory: its active set can change more than native routing's from step to step. Our rule gives it one, with a single parameter $\lambda$, as a bonus for the previous step's experts:

**Algorithm 1:** Persistence-aware OEA, for one layer at decode step $t$

**Input:** router logits $g_i(e)$ for each token $i$; previous set $S_{t-1}$ (empty at the first step); $k_0$, $k$, $\lambda$
1: Selection scores: $s_i(e) = g_i(e) + \lambda \cdot \mathbf{1}[e \in S_{t-1}]$
2: Each token keeps its top-$k_0$ experts by $s_i$; $S_t \leftarrow$ the union over the batch
3: Each token scans its remaining ranking and adds experts already in $S_t$ until it holds $k$
4: Gate weights: the *original* router probabilities of the chosen experts, renormalised. $\lambda = 0$ reproduces OEA exactly

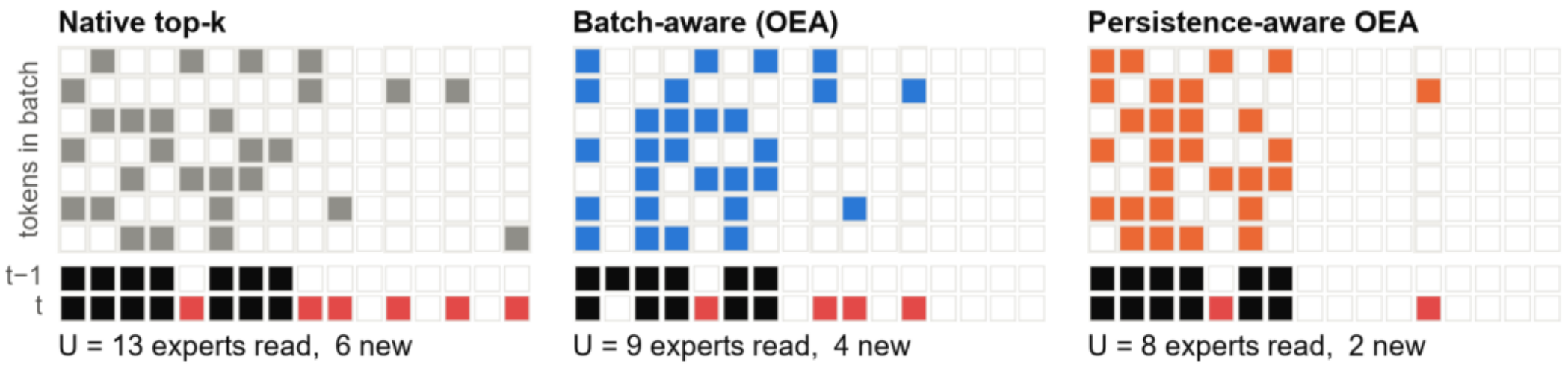

Figure 1: The rules on a toy batch, for intuition only (synthetic logits, 7 tokens, 16 experts, top-4, $k_0 = 2$, $\lambda = 1$). Coloured cells are the experts each token uses at step $t$; shaded columns are those the batch must read. The strips below mark the experts in use at $t$−1 and $t$, with new experts in red. Native routing reads the most. OEA reads fewer, but still changes set. The persistence bonus keeps the cut and fetches fewer new experts.

# 4 Results

## 4.1 Where the energy goes in vectorised MARL

Energy per environment-step falls 1.8× from 128 to 4,096 environments (Figure 2a), repeating earlier single runs within 6%. Most of that gain is already reached at 1,024: the last quadrupling saves about 12%. At 4,096 environments the PPO update takes 82% of energy, so methods evaluated only on acting address under a fifth of training energy. The GRU and MLP heads take 74.6% of kernel time at 128 environments, where the GPU is idle 45% of the time, and 70.2% at 4,096 (Figure 2b). Equation 2 therefore caps a cheaper network at 3.35× at the larger setting, however good the method. These are time shares rather than energy shares, a limit we return to in Section 5.

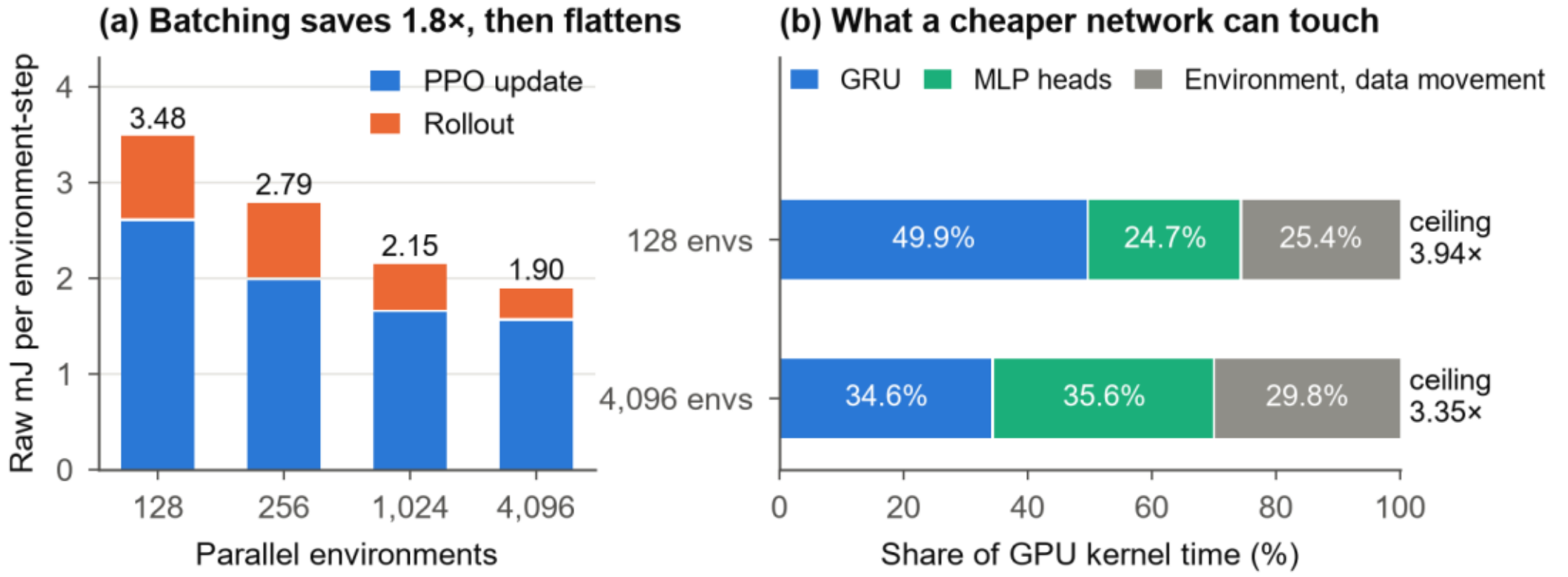

Figure 2: Energy in IPPO on smacv2_10_units. (a) Raw energy per environment-step, mean of 3 seeds. (b) Share of GPU kernel time, with the Amdahl ceiling for an arbitrarily cheap network.

## 4.2 Recurrent policies: structure without persistence, and a control that explains the gains

**Shared patterns exist but do not last.** With $k = 4$ of 16 tiles, 64 signatures cover 94–100% of rows on 3m and 52–78% on 5m_vs_6m, far above random nulls (6–7% within rows, and 21–66% for a stronger across-row shuffle). Yet 6 of 7 trained policies keep fewer than a quarter of rows on the same signature from one step to the next (Figure 3a), and on 5m_vs_6m training does not concentrate activity beyond an untrained network. Persistence does not follow skill either: the best-playing 3m seed has that map's lowest persistence, and the 92% policy on 5m_vs_6m is the least concentrated of all ($D = 951$ distinct signatures, Appendix C).

**Masking a frozen policy fails.** Updating only $k$ tiles of a policy not trained for it costs 14 points of win rate even at $k = 8$ on 3m, and almost everything on the 92% policy on 5m_vs_6m when the mask must be chosen causally, before the update (Figure 3b). Sharing is not the obstacle: snapping rows to 64 shared masks costs at most about 2 points on 3m and about 1 point on 5m_vs_6m against each row's own mask. Skipping is what costs.

**Trained-in states match dense policies, but do not need to switch.** Training with a codebook of $K = 8$ or 32 shared six-tile masks and a Gumbel-softmax selector [33] matches dense 2,048 on 3m within 1.5% of return and beats dense 832 by 1.6–6.6% on 5m_vs_6m. A control then removes the selector: one fixed mask, used at every step, does as well on both maps (90.8% against 89.9% on 3m, 33.1% against 33.3% on 5m_vs_6m). Selector collapse occurred in 4 of 27 gated seeds on 5m_vs_6m, and switching added nothing measurable.

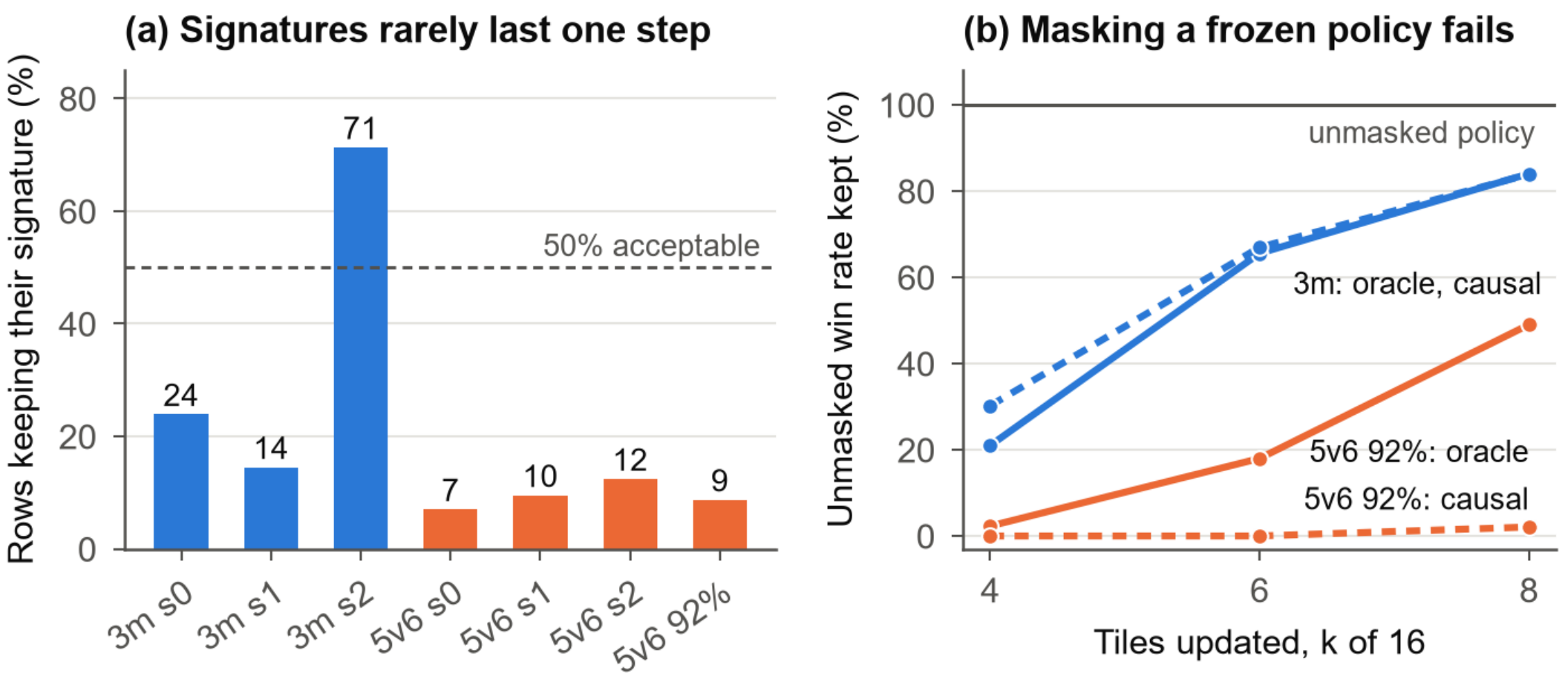

Figure 3: (a) Share of rows keeping their top-4-of-16 signature for one step; 3m in blue, 5m_vs_6m in orange. (b) Win rate kept, relative to the unmasked policy, when only $k$ tiles update. Oracle masks (solid) need the full update; causal masks (dashed) do not.

**A fixed mask is a narrow GRU.** Frozen tiles start at zero and reset to zero, so under a fixed mask they hold exactly zero throughout. The network therefore computes the same function as a 768-unit GRU between 2,048-wide layers, differing only in initialisation and unused weights. Trained directly, that network matches the fixed mask (33.1%, return 1.173; Figure 4a), while plain dense 768 sits with dense 832, so the gap comes from the wide layers rather than from masking or switching. Masking counted only by its recurrent multiply (14% of dense 2,048) also looks cheap, but measured over whole training steps on 3m the masked networks cost 4× dense 832 and 11× dense 416, because the mask is applied after a full 2,048-wide update (Figure 4b). A kernel that skipped frozen tiles would remove part of that cost, though not the wide layers. Only one of our 5m_vs_6m seeds, a dense 2,048 policy, played that map well (92%), a level that no gated or fixed-mask seed approached.

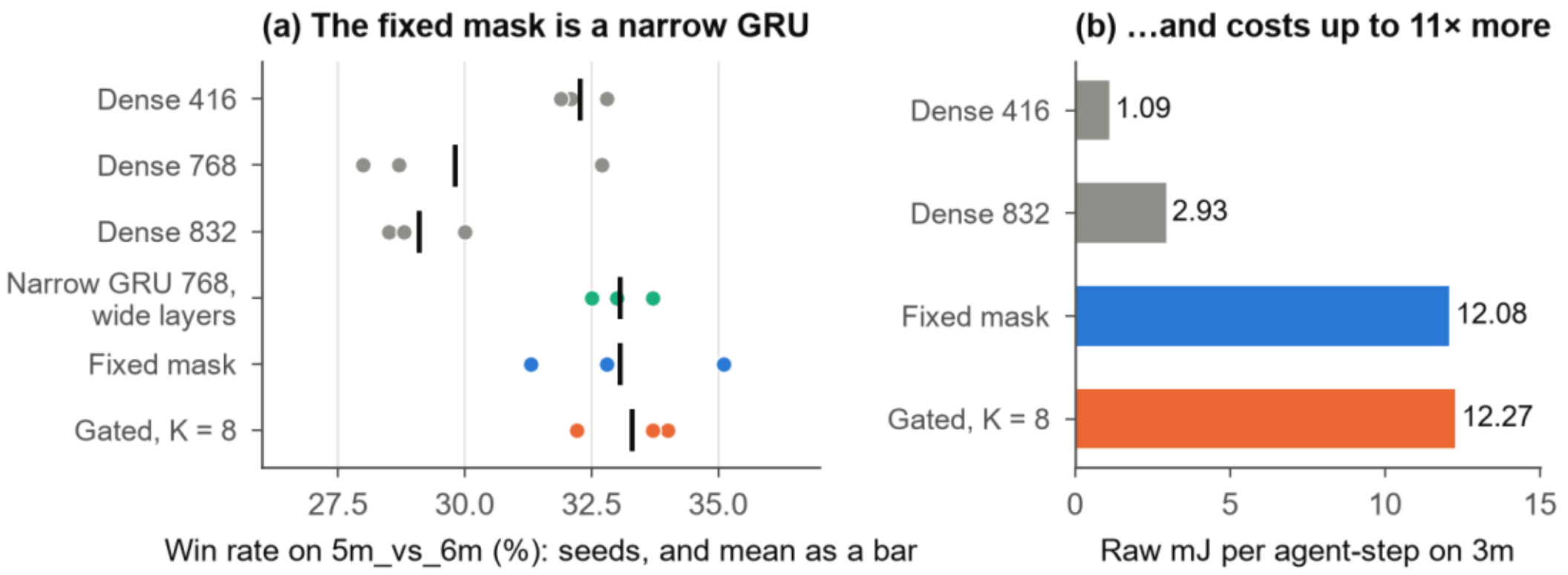

Figure 4: (a) Win rate on 5m_vs_6m, 3 seeds, 50M steps. Dense 2,048 is off this scale (26.9, 29.8 and 92.0%). (b) Measured energy per agent-step over whole training steps on 3m, with clocks locked. The two panels use different maps.

### 4.3 Test-time steering adds nothing beyond re-picking

On frozen 3m policies at $k = 6$, steering the hidden state without a warm-up does not help (42% wins against 43% unsteered). With an added 8-step warm-up, gradient steering reaches 88.6% wins at KL 0.017, but only by writing back 82% of the skipped update, touching every tile and costing 615% of a policy step. Restricted to active tiles, the same setting falls to 56.8%, below unsteered masking (72.4%). Since a run-time method would need to avoid that work, this is no saving. On OLMoE routing, where nothing is skipped and steering has its best chance, it still loses to direct re-picking: within KL 0.05 at $B = 64$, the best steering reaches 0.968 $U_0$ against 0.877 for R-margin. Router-only steering matches the full method, so moving expert inputs adds nothing beyond the swaps a router could make directly.

### 4.4 Routing: granularity sets the room to cut

Native routing on OLMoE uses 77–96% of its experts, so little can be cut (Figure 5a). On Qwen3 a batch uses only 39–60%, in line with OEA's own count at $B = 16$ (49.4 against 48.8 experts). Within KL 0.05, OEA gets furthest at every batch size, cutting $U$ by 25%, 28% and 31% (Figure 5b), which at $B = 64$ means 23.9 rather than 34.7 GB of expert weights read per step. R-margin cuts 22–24%, about twice its 12% on OLMoE at $B = 64$, so most of that earlier limit reflected saturation and not a limit of re-picking. No method reaches 1.5× within the budget, though OEA comes close at 1.45×. Top-1 agreement stays at 0.92–0.93, and benchmark changes are within noise for 500 items, the largest being −2.8 points on unnormalised ARC-Easy (standard error of a difference about 2.6). All rules also make load more even.

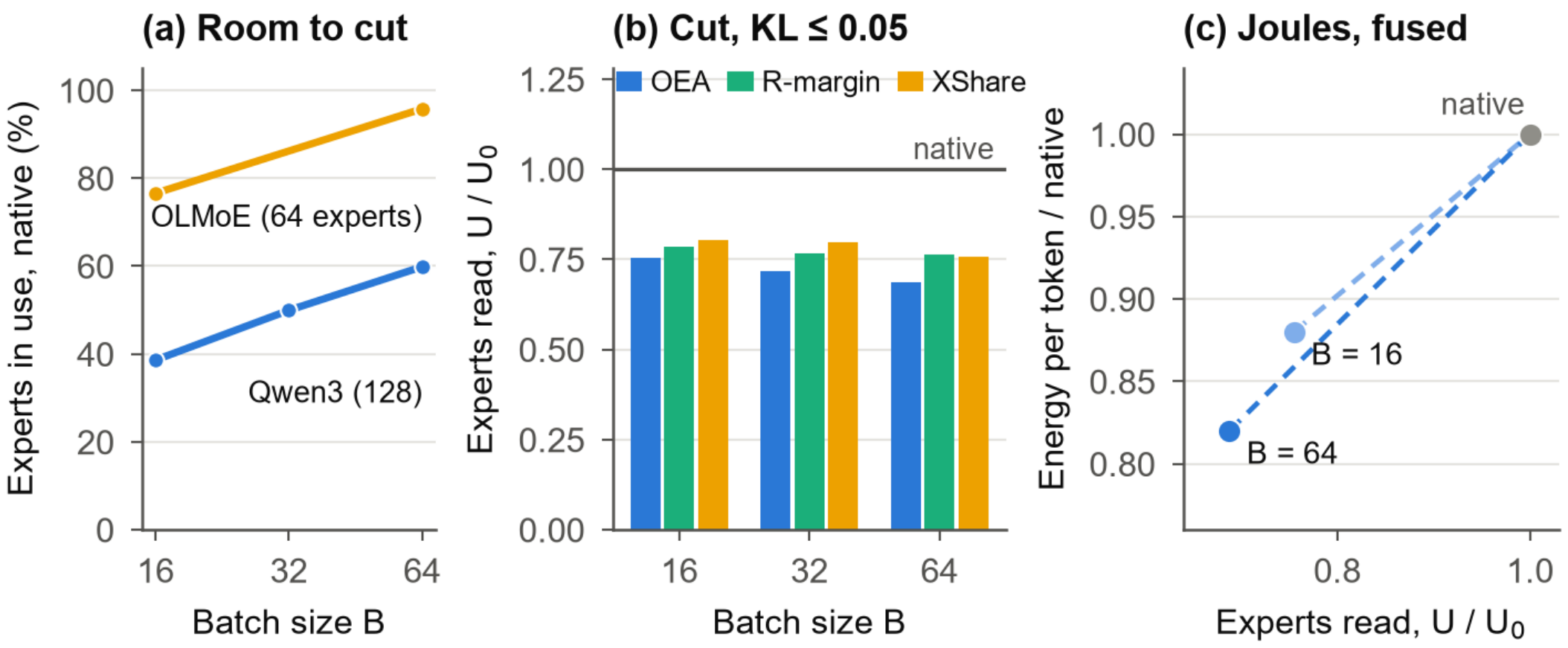

Figure 5: (a) Share of experts a batch uses under native routing. (b) Lowest $U/U_0$ of each rule on Qwen3 within KL 0.05, mean of 3 shards. (c) Energy per token against expert count for OEA on the fused forward; the dashed lines are straight-line fits through two points, used for the extrapolation in Section 5.

### 4.5 Routing in joules

With clocks locked and native routing measured around each rule, OEA uses 16–18% less energy per token on the library's per-expert loop. To remove Python overhead we wrote a fused forward that gathers only the active experts' weights and runs three batched matrix products per layer. It is 4–5× faster and uses 2.1–2.4× less energy than the loop. The saving survives: **12% at $B = 16$ and 18% at $B = 64$** (Table 3), with spread across rounds under 1%. The saving is smaller than the 25–31% cut in weights read, because attention, the router, embeddings and fixed per-step costs do not shrink. Mean power on the fused forward stays within 365–386 W for the two cards (our calculation from Appendix D), so energy largely follows time, falling somewhat faster than the step time (7% and 16%). The rule itself costs 3.4–5.0% of the fused step, and it was not optimised.

### 4.6 Keeping the cut and the persistence

OEA's lead has a cost on the time axis: at $B = 32$ and 64 it needs *more* new experts per step than native routing (9.0 against 7.1 at $B = 64$). Persistence-aware OEA removes that conflict (Figure 6, Table 3). At $B = 32$ and 64 it matches OEA's cut with about 40% fewer new experts than native routing and a higher Jaccard overlap. At $B = 16$ it cuts further than OEA (0.711 against 0.755) with 24% fewer new experts, though its Jaccard (0.631) stays just below native (0.649). Larger $\lambda$ trades cut for persistence smoothly, until KL leaves the budget at $\lambda = 1$–2. The $B = 64$ setting sits at the budget's edge (KL 0.049, and 0.0512 on one shard); a safer setting ($k_0 = 5$, $\lambda = 0.5$) gives 0.759 $U_0$ at KL 0.029 with 4.2 new experts. Top-1 agreement is 0.914 against OEA's 0.920. We have run neither energy nor benchmarks on this rule.

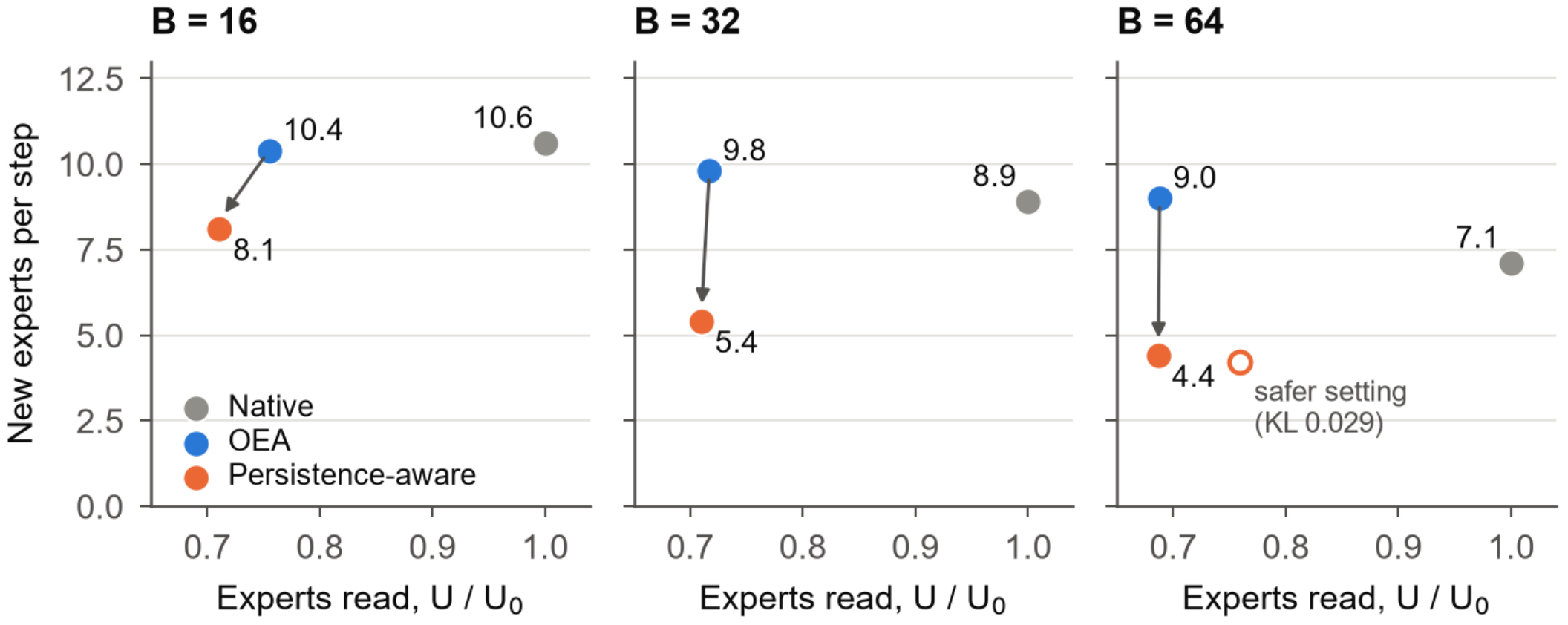

Figure 6: Persistence-aware OEA on Qwen3, best settings within KL 0.05, mean of 3 shards. Down and to the left is better: the rule keeps OEA's cut while fetching fewer new experts than native routing.

Table 3: Routing on Qwen3-30B-A3B, three shards. Energy is on the fused forward, shard 0. "n.m." means not measured.

| $B$ | Rule (setting) | $U$ | $U/U_0$ | KL | Jaccard | New per step | J/token | vs native |
|---|---|---|---|---|---|---|---|---|
| 16 | Native | 49.4 | 1.000 | 0 | 0.649 | 10.6 | 3.79 | 1.00 |
| 16 | OEA ($k_0 = 5$) | 37.3 | 0.755 | 0.032 | 0.571 | 10.4 | 3.33 | 0.88 |
| 16 | Persistence-aware ($k_0 = 5, \lambda = 0.25$) | 35.1 | 0.711 | 0.043 | 0.631 | 8.1 | n.m. | n.m. |
| 32 | Native | 63.9 | 1.000 | 0 | 0.757 | 8.9 | n.m. | n.m. |
| 32 | OEA ($k_0 = 4$) | 45.8 | 0.717 | 0.037 | 0.652 | 9.8 | n.m. | n.m. |
| 32 | Persistence-aware ($k_0 = 5, \lambda = 0.5$) | 45.4 | 0.710 | 0.046 | 0.789 | 5.4 | n.m. | n.m. |
| 64 | Native | 76.6 | 1.000 | 0 | 0.830 | 7.1 | 1.54 | 1.00 |
| 64 | OEA ($k_0 = 3$) | 52.7 | 0.688 | 0.045 | 0.710 | 9.0 | 1.27 | 0.82 |
| 64 | Persistence-aware ($k_0 = 4, \lambda = 0.5$) | 52.6 | 0.687 | 0.049 | 0.847 | 4.4 | n.m. | n.m. |

## 5 Discussion

**Why routing works where masking did not.** An expert is a complete computation. Re-routing a token changes which complete computations contribute to it and skips nothing inside any of them, so its effect is bounded and can be measured with KL. A frozen tile, by contrast, skips part of an update that the rest of the state depends on, so a frozen policy loses quality quickly, and the only steering that restores it is the skipped update itself. Discrete decisions at a router look like a much better place to spend a quality budget than partial updates of a dense state.

**A saving must beat the right control.** Three of our apparent gains weakened once tested against a simpler control: gating against a fixed mask, a fixed mask against a narrow GRU, and recurrent-multiply share against measured joules. We suggest that any claim for conditional computation should report whole-step joules, and should beat both a fixed mask and the smallest dense network that matches return. Our SMAX maps have identical units and short episodes (15–19 steps), so these results do not rule out switching on tasks with distinct phases.

**How far routing can go.** A straight line through the native and OEA points on the fused forward (Figure 5c) leaves about 42% of energy at $B = 64$, and 51% at $B = 16$, outside whatever follows expert count. If that held, routing alone would be capped at roughly 2–2.4×. This is our own extrapolation from two points per batch size and one implementation, so it is only indicative. It does imply that further gains need the remaining energy to be addressed as well, not only lower $U$.

**What persistence is worth remains open.** Published gains from persistence come mainly from avoiding transfers of offloaded weights [18, 22]. Our weights stay in GPU memory and our fused forward does not exploit stable experts, so we have shown that persistence can be obtained cheaply but not yet that it saves joules on this hardware. Training a router against a joint batch-and-time objective, perhaps with a deliberation cost [21], is a natural next step, and our rule is a free baseline such a method would have to beat.

**Limitations.** The results come from one GPU type, one fine-grained model and one text source, with consecutive windows as shards, so the close agreement between shards (within 0.016 in $U/U_0$) understates variation on other text. Decoding is teacher-forced, so errors cannot compound as they would in free generation. Benchmarks ran at $B = 16$ only, on 500 items, with a standard error of about 2 points, so they can exclude only large losses. MoE energy was measured on shard 0, on board counters that exclude CPU and host memory; a fast external meter [34] would be a useful cross-check. Idle power drifted during the session (126 to 81 W for the two cards), which leaves absolute above-idle values uncertain by about ±20% but leaves relative savings essentially unaffected. MARL kernel shares are time, not energy, and many MARL comparisons rest on 3–5 seeds. Our literature statements come from targeted rather than systematic searches.

# 6 Conclusion

Sharing computation across a batch is measurably worth doing, but where it works depends on the kind of decision being shared. On recurrent policies, activity patterns did not persist, frozen policies did not tolerate masking, and trained-in masks reduced to a narrow GRU with no measured saving. On a fine-grained MoE model, batch-aware routing saved 12–18% of the energy per token within a tight quality budget, and a one-parameter rule kept that cut in expert count while fetching about 40% fewer new experts than native routing at larger batches. The next measurements are the energy and benchmark quality of that rule, free generation, and a kernel that exploits stable experts, which together will show how much of the remaining energy routing can reach.

**Data and code.** Scripts and results files are available from the authors upon request.
**Acknowledgements.** This work was supported by the Engineering and Physical Sciences Research Council in the UK (EPSRC) under grant EP/T000783/1 and the Economic and Social Research Council in the UK (ESRC) under grant ES/Z00036X/2. The AI assistant Claude (Anthropic) was used to help write and review code, draft parts of the text, and language editing. It did not generate experimental results or observations. The authors checked the code, results and text, and take full responsibility for the paper's content.

# A Protocols

**Hardware and software.** NVIDIA RTX A6000 (Ampere, 48 GB, 300 W limit). MARL structure runs: driver 555.42.06, JAX 0.10.2, Flax 0.12.8. Qwen3 runs: transformers 4.51.3, bf16 weights split across two cards.

**MARL energy.** IPPO with a GRU of 128 units on smacv2_10_units for 10M steps, with 128, 256, 1,024 and 4,096 environments and 3 seeds per setting. Seeds of one setting ran at the same time, one per GPU; the original single runs ran one at a time. Energy is NVML over rollout and update, with a 60 s idle baseline. Values are exported directly from experiment logs. Two PPO updates were profiled per seed.

**Recurrent structure.** Training used 256 environments; analysis rollouts used 200 steps on 4,096 environments (12,288 agent-steps per step on 3m, 20,480 on 5m_vs_6m), excluding dead agents. 3m ran for 10M steps, and 5m_vs_6m for 50M because at 10M every dense width was still improving. Testing used 20,480 first episodes per condition, with a standard error of about 0.3 points. Causal masks use the previous hidden state; oracle masks use the candidate new state. Gated policies use straight-through Gumbel-softmax, with noise stored in the rollout and replayed in the update, and test with argmax. With hysteresis margin $\delta$, the current state is kept unless another scores higher by more than $\delta$. The 3m energy runs were 2M steps, one seed each, 3 interleaved rounds, with a clock locked at 1,800 MHz: long enough for steady-state cost, not for final performance.

**MoE routing.** WikiText validation cut into windows (a 128-token prompt and 128 teacher-forced continuation positions), with three shards of 256 passages. The tokens of a batch at one position form one decode step. With no rule, the replacement forward reproduces the model's logits exactly. Weights read per step are layers × $U$ × expert size (9.4 MB on Qwen3, 12.6 MB on OLMoE), a memory-traffic model and not an energy measurement. XShare was published on a different model and its paper does not state whether gate weights are renormalised after restriction; we renormalise, as Qwen3 does. OEA can leave a token with fewer than $k$ experts; this did not occur in our OEA runs.

**MoE energy.** Qwen3 on two cards, graphics clocks locked at 1,800 MHz. Each batch's prompt is prefilled without measurement, then 128 continuation tokens are decoded with the rule active. NVML is summed over both cards and read around the decode loops only. There were 3 rounds of the rules, with native before and after each, and at least 5 minutes per condition on shard 0. On one MoE block the fused forward matches the library loop to 0.7% relative error. It is our own, written for this run, and is not ScatterMoE.

**Persistence-aware sweep.** $\lambda \in \{0, 0.25, 0.5, 1, 2\} \times k_0 \in \{3, 4, 5, 6\}$ on shard 0 at each batch size, with chosen settings repeated on shards 1 and 2. In a targeted, non-systematic search we found no earlier rule of this exact form. ReMoE [18] biases the router towards recent experts by fine-tuning, acting per token rather than across a batch.

# B MARL energy and kernel time

Table 4: Energy per environment-step on smacv2_10_units (mJ, mean ± sd over 3 seeds).

| Environments | Raw total | Raw update | Raw rollout | Above idle | Original runs, raw total |
|---|---|---|---|---|---|
| 128 | 3.48 ± 0.14 | 2.59 ± 0.10 | 0.90 ± 0.05 | 3.06 ± 0.16 | 3.35, 3.30 |
| 256 | 2.79 ± 0.02 | 1.98 ± 0.02 | 0.80 ± 0.01 | 2.43 ± 0.04 | 2.67, 2.81 |
| 1,024 | 2.15 ± 0.02 | 1.64 ± 0.04 | 0.51 ± 0.02 | 1.94 ± 0.03 | 2.21, 2.12 |
| 4,096 | 1.90 ± 0.04 | 1.55 ± 0.03 | 0.34 ± 0.01 | 1.72 ± 0.05 | 1.92 |

Table 5: Share of GPU kernel time in the network (GRU and MLP heads, rollout and update), 3 seeds.

| Environments | Network share | Ceiling | GRU alone | GPU busy |
|---|---|---|---|---|
| 128 | 74.6% ± 0.07% | 3.94× | 49.9% | 55% |
| 4,096 | 70.2% ± 0.04% | 3.35× | 34.6% | 97% |

**First-run profile.** A single earlier profile at 128 environments gives the finer split in Table 6 (exact values from the profile's outputs). It shows the structure of an iteration rather than final shares. In this profile, environment simulation and copies are 87% of rollout time, so a cheaper policy barely helps acting.

Table 6: Kernel time per iteration, first single-run profile.

| Component | ms | Share of phase |
|---|---|---|
| **PPO update** | **231.1** | **100%** |
| GRU recurrent matmuls | 52.15 | 22.6% |
| GRU input projections | 67.46 | 29.2% |
| GRU gates and carry reset | 17.58 | 7.6% |
| MLP actor-critic heads | 66.40 | 28.7% |
| PPO loss, optimiser and GAE | 1.74 | 0.8% |
| Shuffling, minibatching, loop | 25.75 | 11.1% |
| **Rollout** | **50.5** | **100%** |
| GRU (matmuls, projections, gates) | 2.72 | 5.4% |
| MLP actor-critic heads | 3.70 | 7.3% |
| Environment simulation, copies, other | 44.08 | 87.3% |

# C Recurrent policies: full tables

Table 7: Shared structure at $k = 4$ of 16 tiles. $D$ is distinct signatures per step (1,820 possible). Rollout win rate in brackets.

| Policy | *D* | Coverage | Persistence |
|---|---|---|---|
| 3m seed 0 (87%) | 77.3 | 99.8% @64 | 23.9% |
| 3m seed 1 (91%) | 186.5 | 48.7% @8 | 14.5% |
| 3m seed 2 (90%) | 34.9 | 97.5% @8 | 71.3% |
| 3m untrained | 353–403 | 87.8% @64 | 27–28% |
| 5m_vs_6m seed 0 (32%) | 605.0 | 69.2% @64 | 7.0% |
| 5m_vs_6m seed 1 (26%) | 691.5 | 72.0% @64 | 9.5% |
| 5m_vs_6m seed 2 (25%) | 509.4 | 78.3% @64 | 12.4% |
| 5m_vs_6m 92% policy | 950.7 | 52.1% @64 | 8.7% |
| 5m_vs_6m untrained (one network) | 358 | 90% @64 | 27% |
| Null: tiles permuted within row | ≈1,814–1,820 | 6–7% @64 | ≤ 0.1% |

At $k = 6$ (8,008 possible signatures) the picture is the same: on 3m, $D$ is 51–415 and persistence 11–84%; on 5m_vs_6m, $D$ is 998–2,300, coverage@64 34–63% and persistence 6–8%. The one passing seed (3m seed 2) looks tile-dominated, since eight signatures cover 97.5% of rows and it has the highest across-row-null coverage (66%). Hysteresis with warm-up on masked frozen policies capped persistence at about 50–66% without raising the win rate.

Table 8: Win rate when only $k$ of 16 tiles update, no warm-up unless stated. Unmasked: 87.7% (3m), 33.6% (5m_vs_6m seed 0), 91.2% (92% policy).

| Mask | 3m: $k = 4$ | 6 | 8 | Seed 0: $k = 4$ | 6 | 8 | 92%: $k = 4$ | 6 | 8 |
|---|---|---|---|---|---|---|---|---|---|
| Own, oracle | 18.4 | 57.5 | 73.7 | 0.7 | 3.8 | 10.2 | 2.1 | 16.4 | 44.7 |
| Shared 64, oracle | 18.5 | 57.7 | 73.9 | 0.6 | 4.2 | 11.4 | 1.7 | 15.8 | 52.8 |
| Random tiles | 36.0 | 57.5 | 70.2 | 0.4 | 1.9 | 5.8 | 0.0 | 0.0 | 1.0 |
| Shared 64, causal | 26.5 | 58.8 | 73.7 | 0.0 | 1.1 | 8.1 | 0.0 | 0.0 | 1.9 |
| Shared 64, causal, warm-up 8 | – | – | – | 9.1 | 10.8 | 13.5 | 0.0 | 0.0 | 0.0 |
| Random, warm-up 8 | 77.8 | 83.4 | 85.4 | 8.3 | 9.5 | 14.0 | 0.0 | 0.2 | 9.5 |

Table 9: State-gated training, 3 seeds per row. Dwell on 5m_vs_6m is the median, because a seed stuck in one state dominates a mean. Active recurrent compute: dense 2,048 100%, 832 16.5%, 416 4%, gated 14%. * One dense 2,048 seed reached 92%; it is reported separately rather than averaged.

| Policy | 3m return | 3m win | Persist. | Dwell | 5v6 return | 5v6 win | Persist. | Dwell |
|---|---|---|---|---|---|---|---|---|
| Dense 2,048 | 1.884 | 89.7% | | | 1.082, 1.126, 1.902* | 27–30, 92%* | | |
| Dense 832 | 1.907 | 91.7% | | | 1.116 | 29.1% | | |
| Dense 416 | 1.911 | 92.1% | | | 1.160 | 32.3% | | |
| $K = 8$, $\delta = 0$ | 1.886 | 89.9% | 42% | 1.8 | 1.174 | 33.3% | 63% | 2.5 |
| $K = 8$, $\delta = 0.5$ | 1.889 | 90.2% | 74% | 3.9 | 1.171 | 33.0% | 89% | 34.0 |
| $K = 8$, $\delta = 1$ | 1.885 | 89.8% | 87% | 9.3 | 1.190 | 34.4% | 71% | 2.4 |
| $K = 32$, $\delta = 0$ | 1.872 | 88.7% | 45% | 1.8 | 1.167 | 32.8% | 42% | 1.7 |
| $K = 32$, $\delta = 0.5$ | 1.869 | 88.4% | 71% | 3.4 | 1.134 | 30.5% | 60% | 2.7 |
| $K = 32$, $\delta = 1$ | 1.870 | 88.5% | 87% | 7.9 | 1.161 | 32.3% | 73% | 2.7 |
| $K = 32$ random, $\delta = 0$ | 1.857 | 87.4% | 42% | 1.7 | 1.150 | 31.5% | 80% | 3.9 |
| $K = 32$ random, $\delta = 0.5$ | 1.861 | 87.7% | 67% | 3.1 | 1.148 | 31.4% | 71% | 3.6 |
| $K = 32$ random, $\delta = 1$ | 1.856 | 87.3% | 80% | 5.0 | 1.135 | 30.7% | 63% | 2.4 |

Table 10: Fixed-mask and narrow-GRU controls, win rate per seed. On 5m_vs_6m, 5 seeds did not fit in one GPU's memory.

| Map | Condition | Seeds | Win rate per seed (%) | Mean win | Mean return |
|---|---|---|---|---|---|
| 3m | Fixed mask | 5 | 88.2, 91.1, 91.7, 91.6, 91.4 | 90.8% | 1.896 |
| 3m | Gated $K = 8$, $\delta = 0$ | 5 | 90.0, 89.8, 89.9, 89.8, 89.9 | 89.9% | 1.886 |
| 3m | Dense 416 | 5 | 91.3, 91.8, 93.1, 92.9, 92.7 | 92.4% | 1.914 |
| 3m | Dense 832 | 3 | 91.8, 92.5, 91.0 | 91.7% | 1.907 |
| 3m | Dense 2,048 | 3 | 87.4, 91.8, 89.9 | 89.7% | 1.884 |
| 5v6 | Narrow GRU 768, wide layers | 3 | 32.5, 33.7, 33.0 | 33.1% | 1.173 |
| 5v6 | Fixed mask | 3 | 31.3, 32.8, 35.1 | 33.1% | 1.173 |
| 5v6 | Gated $K = 8$, $\delta = 0$ | 3 | 34.0, 33.7, 32.2 | 33.3% | 1.175 |
| 5v6 | Gated $K = 8$, $\delta = 1$ | 3 | 33.5, 36.3, 33.3 | 34.4% | 1.190 |
| 5v6 | Dense 416 | 3 | 32.1, 31.9, 32.8 | 32.3% | 1.160 |
| 5v6 | Dense 768 | 3 | 28.7, 28.0, 32.7 | 29.8% | 1.127 |
| 5v6 | Dense 832 | 3 | 28.5, 28.8, 30.0 | 29.1% | 1.116 |
| 5v6 | Dense 2,048 | 3 | 26.9, 29.8, 92.0 | 49.6% | 1.370 |

Table 11: Measured compute on 3m: energy per agent-step over whole training steps, 2M-step runs, clock locked at 1,800 MHz.

| Condition | Raw mJ | Above idle | Training time | Against dense 832 |
|---|---|---|---|---|
| Dense 416 | 1.09 ± 0.01 | 0.76 ± 0.01 | 32 s | 0.37× |
| Dense 832 | 2.93 ± 0.01 | 2.27 ± 0.01 | 62 s | 1.00× |
| Fixed mask | 12.08 ± 0.02 | 9.44 ± 0.02 | 247 s | 4.1× |
| Gated $K = 8$, $\delta = 0$ | 12.27 ± 0.04 | 9.54 ± 0.05 | 251 s | 4.2× |

The energy of the narrow GRU itself has not been measured. It should cost about the wide layers plus a 768-unit GRU, well below the fixed mask.

# D MoE routing: full tables

Table 12: Best setting of each rule on Qwen3 within KL ≤ 0.05, means over three shards. Load is the busiest expert as a multiple of the mean.

| *B* | Rule (setting) | *U/U*₀ [range] | KL | Top-1 | Per-plex. | Jaccard | Load | Read |
|---|---|---|---|---|---|---|---|---|
| 16 | Native | 1.000 | 0 | 1.000 | 10.73 | 0.65 | 4.2× | 22.4 GB |
| 16 | R-margin ($N = 32$, $\tau = 0.75$) | 0.784 [0.775–0.790] | 0.038 | 0.923 | 10.98 | 0.63 | 3.6× | 17.5 GB |
| 16 | OEA ($k_0 = 5$) | 0.755 [0.753–0.757] | 0.032 | 0.928 | 10.87 | 0.57 | 3.4× | 16.9 GB |
| 16 | XShare ($k_0 = 2$, budget 24) | 0.803 [0.792–0.808] | 0.048 | 0.919 | 11.03 | 0.64 | 3.6× | 18.0 GB |
| 32 | Native | 1.000 | 0 | 1.000 | 10.74 | 0.76 | 5.0× | 29.0 GB |
| 32 | R-margin ($N = 32$, $\tau = 0.75$) | 0.768 [0.761–0.774] | 0.043 | 0.919 | 10.99 | 0.70 | 4.3× | 22.3 GB |
| 32 | OEA ($k_0 = 4$) | 0.717 [0.715–0.720] | 0.037 | 0.925 | 10.89 | 0.65 | 3.9× | 20.8 GB |
| 32 | XShare ($k_0 = 2$, budget 24) | 0.796 [0.786–0.802] | 0.041 | 0.927 | 10.95 | 0.74 | 4.2× | 23.1 GB |
| 64 | Native | 1.000 | 0 | 1.000 | 10.73 | 0.83 | 5.7× | 34.7 GB |
| 64 | R-margin ($N = 48$, $\tau = 1$) | 0.763 [0.754–0.768] | 0.049 | 0.916 | 11.02 | 0.79 | 4.6× | 26.5 GB |
| 64 | OEA ($k_0 = 3$) | 0.688 [0.686–0.690] | 0.045 | 0.920 | 10.97 | 0.71 | 4.2× | 23.9 GB |
| 64 | XShare ($k_0 = 2$, budget 16) | 0.758 [0.753–0.762] | 0.040 | 0.926 | 10.93 | 0.79 | 4.5× | 26.3 GB |

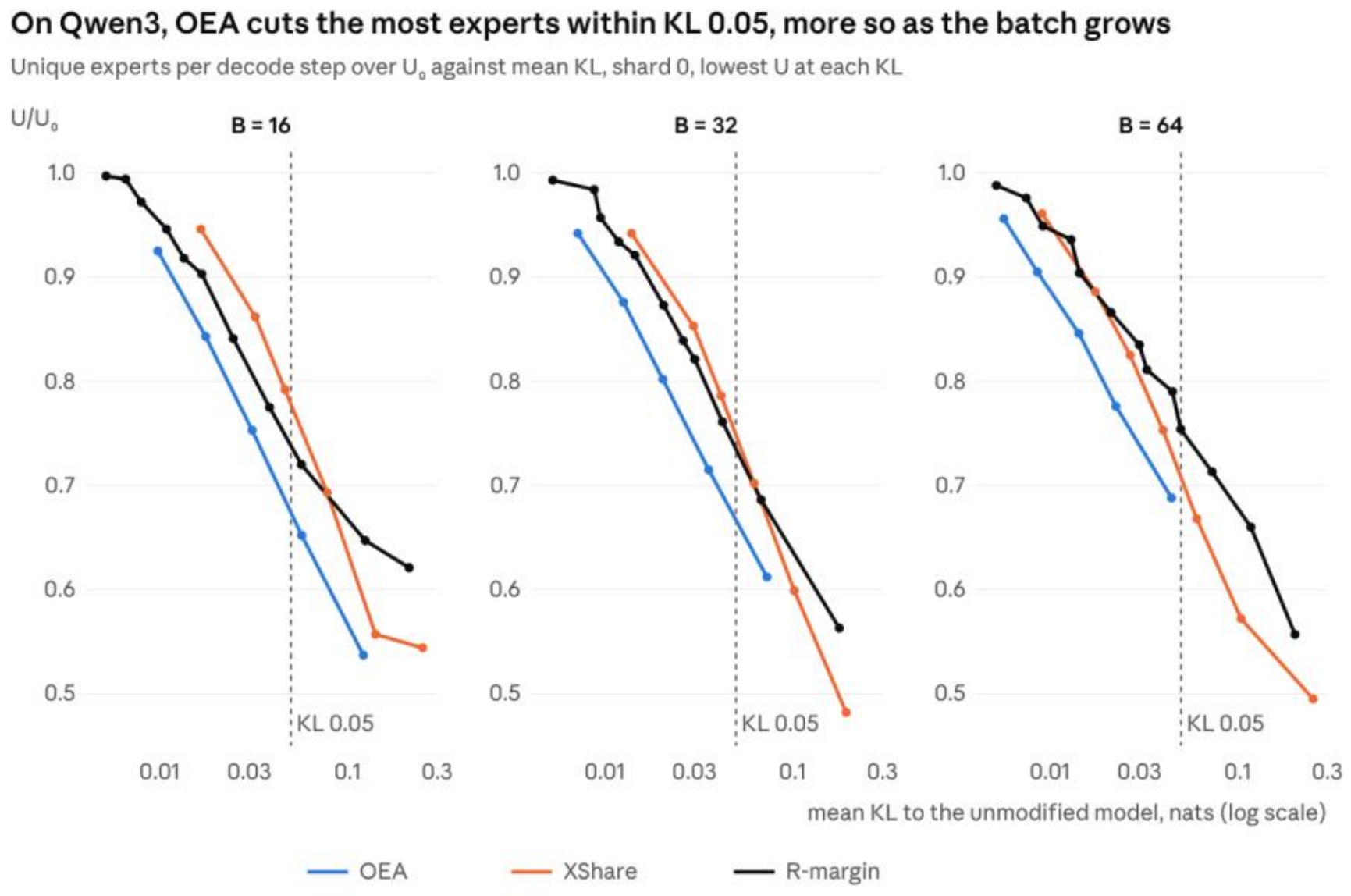

Figure 7: Unique experts per decode step against mean KL on Qwen3-30B-A3B, shard 0, with the lowest $U$ at each KL over each rule's grid. Beyond the budget all three rules keep falling, to about 0.55–0.65 $U_0$ by KL 0.1–0.2.

Table 13: Energy per decoded token on Qwen3, shard 0, mean ± sd over rounds. "Against native" divides each measurement by the mean of the native measurements just before and after it.

| Forward | *B* | Rule | Raw J/token | Above idle | Against native | ms per step |
|---|---|---|---|---|---|---|
| Library | 16 | Native | 8.95 ± 0.19 | 3.87 ± 0.06 | 1.00 | 783 ± 22 |
| Library | 16 | OEA | 7.50 ± 0.22 | 3.22 ± 0.10 | 0.84 [0.81–0.86] | 659 ± 18 |
| Library | 16 | R-margin | 7.78 ± 0.34 | 3.34 ± 0.14 | 0.87 [0.84–0.91] | 684 ± 31 |
| Library | 16 | XShare | 7.91 ± 0.15 | 3.40 ± 0.08 | 0.89 [0.86–0.90] | 696 ± 11 |
| Library | 64 | Native | 3.29 ± 0.06 | 1.51 ± 0.02 | 1.00 | 1,098 ± 22 |
| Library | 64 | OEA | 2.67 ± 0.06 | 1.24 ± 0.03 | 0.82 [0.80–0.84] | 886 ± 19 |
| Library | 64 | R-margin | 2.85 ± 0.01 | 1.32 ± 0.00 | 0.87 [0.86–0.88] | 947 ± 2 |
| Library | 64 | XShare | 2.83 ± 0.04 | 1.31 ± 0.01 | 0.87 [0.86–0.87] | 941 ± 15 |
| Fused | 16 | Native | 3.79 ± 0.01 | 2.69 ± 0.01 | 1.00 | 157 ± 1 |
| Fused | 16 | OEA | 3.33 ± 0.01 | 2.32 ± 0.01 | 0.88 [0.88–0.88] | 146 ± 1 |
| Fused | 64 | Native | 1.54 ± 0.01 | 1.09 ± 0.01 | 1.00 | 256 ± 1 |
| Fused | 64 | OEA | 1.27 ± 0.00 | 0.89 ± 0.00 | 0.82 [0.82–0.83] | 216 ± 0 |

Idle power for the two cards moved from 126 to 81 W during the session (137 and 86 W around the fused run). The routing rules cost 7.5 ms (OEA), 15 ms (R-margin) and 10 ms (XShare) per step, as the median of 200 calls over 48 layers. Mean power in Section 4.5 is raw J/token × $B$ ÷ step time.

Table 14: Benchmark accuracy (%) on Qwen3, 500 items, $B = 16$, change against native in brackets. acc_norm is length-normalised. One figure has a standard error of about 2 points.

| Rule | ARC-Easy acc | ARC-Easy acc_norm | HellaSwag acc | HellaSwag acc_norm |
|---|---|---|---|---|
| Native | 78.2 | 78.6 | 52.4 | 67.2 |
| R-margin | 77.8 (−0.4) | 79.6 (+1.0) | 52.2 (−0.2) | 65.8 (−1.4) |
| OEA | 75.4 (−2.8) | 78.8 (+0.2) | 53.0 (+0.6) | 65.4 (−1.8) |
| XShare | 78.4 (+0.2) | 78.6 (0.0) | 52.2 (−0.2) | 66.0 (−1.2) |

Table 15: Median time per cached decode step (ms), library per-expert loop on two GPUs, best setting at each $B$. These show direction rather than the size of a real speed-up.

| Rule | $B = 16$ | $B = 32$ | $B = 64$ |
|---|---|---|---|
| Native | 761 | 935 | 1,048 |
| R-margin | 670 | 806 | 888 |
| OEA | 651 | 777 | 829 |
| XShare | 668 | 816 | 879 |

# E Steering experiments

**Frozen recurrent policies.** Three frozen dense 2,048 policies on 3m win 87.7, 90.6 and 89.6% unmasked. At each step an agent picks one of 32 shared masks and adds $\delta$ to its previous hidden state. S1 moves towards the mask's mean hidden state; S2 takes 1 or 3 projected gradient steps to make frozen units stationary, with $\|\delta\| \leq 0.5\|h\|$. The 8-step warm-up was added after a first sweep showed it was the only setting where steering moved the win rate, and it reduces the possible saving (62–66% of tile updates run). "Restored" is the share of the skipped update put back by $\delta$.

Table 16: Steering at $k = 6$ on 3m, 3 seeds. "Tiles written" counts a tile if the GRU updates it or $\delta$ changes it.

| Condition | Wins by seed (%) | Mean | KL | Persist. | Tiles written | Re-stored |
|---|---|---|---|---|---|---|
| No warm-up: unsteered | 58.6 / 20.8 / 49.9 | 43.1% | 0.324 | 100% | 42% | 0% |
| Best S1 ($\alpha = 0.05$) | 57.8 / 20.1 / 48.1 | 42.0% | 0.350 | 100% | 100% | 0% |
| Best S2 (3 steps, every step) | 58.1 / 20.0 / 47.7 | 41.9% | 0.188 | 92% | 100% | 27% |
| same, active tiles only | 40.7 / 11.2 / 34.9 | 28.9% | 0.362 | 100% | 42% | 0% |
| Warm-up 8: unsteered | 71.4 / 66.3 / 79.4 | 72.4% | 0.554 | 65% | 66% | 0% |
| Best S1 ($\alpha = 0.4$) | 84.2 / 76.7 / 86.2 | 82.4% | 0.267 | 68% | 100% | 3% |
| same, active tiles only | 73.9 / 68.7 / 79.9 | 74.2% | 0.499 | 56% | 65% | 0% |
| Best S2 (3 steps, every step, no KL term) | 87.5 / 89.5 / 88.9 | 88.6% | 0.017 | 67% | 100% | 82% |
| same, random $\delta$ | 81.4 / 76.7 / 83.0 | 80.4% | 0.313 | 29% | 100% | −66% |
| same, no mask | 87.6 / 89.9 / 89.1 | 88.9% | 0.009 | 66% | 100% | 87% |
| same, active tiles only | 63.5 / 46.9 / 59.9 | 56.8% | 0.631 | 64% | 69% | 0% |

Steering time per call, as a share of one policy step on 12,288 agents, every step: S1 2.5%; S2 with 1 gradient step 169%; S2 with 3 steps 615%. Steered states do not gather on the shared masks: with warm-up, S2 raises distinct signatures to 449 per step against 140 for S1.

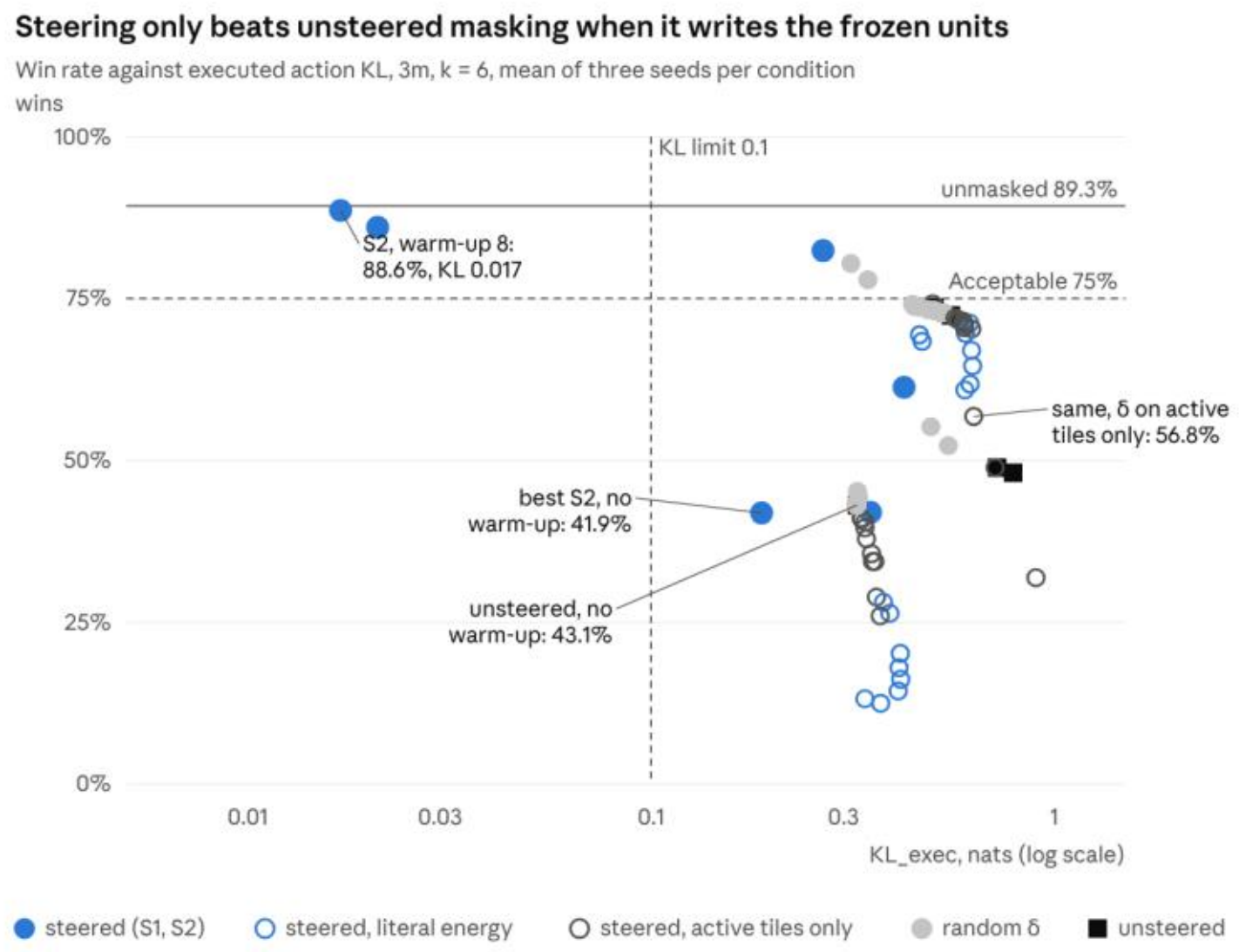

Figure 8: Win rate against executed-action KL at $k = 6$ on 3m, mean of three seeds per condition (71 conditions). Filled points write to frozen units; hollow black points are the same settings restricted to active tiles.

**MoE routing (OLMoE-1B-7B).** The L1 method adds $\delta = \alpha \cdot W^{\mathsf{T}}(q - \mathrm{softmax}(Wx))$ to the block input, clipped to $\beta\|x\|$, and routes normally, where $q$ targets the $N$ experts with the largest summed router probability. Controls are router-only L1, random $\delta$ of equal norm, and $\delta$ with the original top-8 kept. The sweep covered 433 conditions per batch size on shard 0.

Table 17: Best settings on OLMoE, three-shard means. "Same swap" is the share of changed tokens ending with the set R-margin would give them.

| *B* | Condition | $U/U_0$ | KL | Jaccard | Load | Changed | Same swap |
|---|---|---|---|---|---|---|---|
| 64 | Native ($U_0 = 61.3$ of 64) | 1.000 | 0 | 0.960 | 3.40 | – | – |
| 64 | R-margin ($N = 48$, $\tau = 1$) | 0.877 | 0.037 | – | – | – | – |
| 64 | L1 ($N = 24$, $\alpha = 4$, $\beta = 0.05$) | 0.968 | 0.040 | 0.936 | 3.56 | 65% | 52% |
| 64 | same, router only | 0.965 | 0.043 | 0.935 | 3.59 | 65% | 52% |
| 64 | same, random $\delta$ | 1.000 | 0.004 | 0.960 | 3.40 | 8% | – |
| 64 | same, $\delta$ only, original top-8 | 1.003 | 0.009 | 0.961 | 3.37 | 0% | – |
| 64 | L1 + R-margin ($N = 32$) | 0.894 | 0.046 | 0.882 | 3.45 | 77% | 71% |
| 16 | Native ($U_0 = 49.0$ of 64) | 1.000 | 0 | 0.757 | 3.28 | – | – |
| 16 | R-margin ($N = 24$, $\tau = 0.75$) | 0.754 | 0.041 | – | – | – | – |
| 16 | L1 ($N = 32$, $\alpha = 4$, $\beta = 0.1$) | 0.912 | 0.043 | 0.702 | 3.12 | 56% | 39% |
| 16 | same, router only | 0.909 | 0.041 | 0.700 | 3.12 | 55% | 39% |
| 16 | L1 + R-margin ($N = 24$) | 0.780 | 0.047 | 0.630 | 2.97 | 82% | 76% |

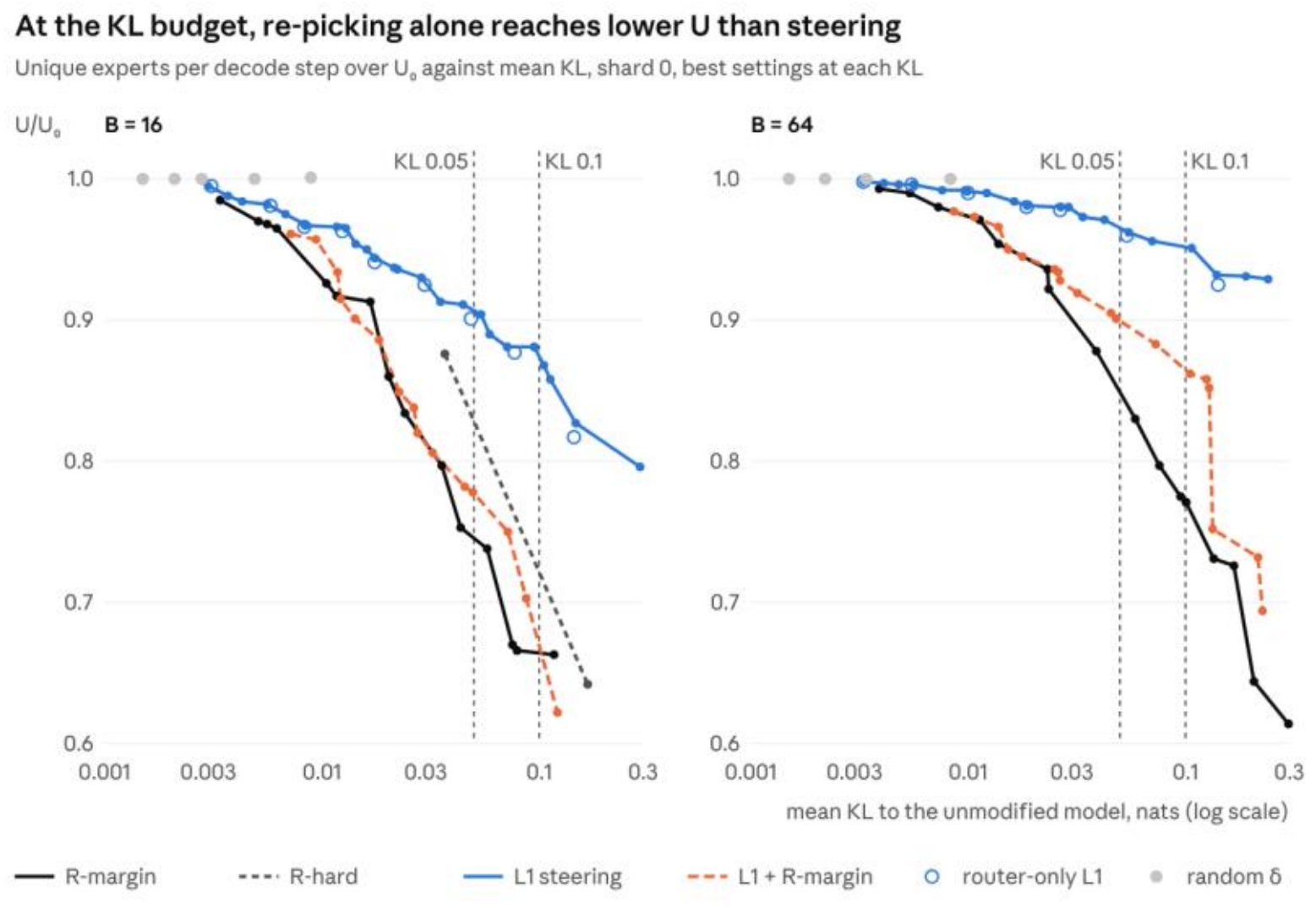

Figure 9: Unique experts per decode step against mean KL on OLMoE, shard 0, best settings at each KL.

Table 18: Benchmark accuracy (%) on OLMoE, 500 items each, change against native at the same batch size in brackets.

| Condition | ARC-Easy, $B = 16$ | ARC-Easy, $B = 64$ | HellaSwag, $B = 16$ | HellaSwag, $B = 64$ |
|---|---|---|---|---|
| Native | 75.8 | 75.6 | 67.4 | 67.0 |
| R-margin ($N = 48$, $\tau = 1$) | 75.6 (−0.2) | 76.2 (+0.6) | 66.8 (−0.6) | 67.2 (+0.2) |
| L1 ($N = 24$) | 76.2 (+0.4) | 76.2 (+0.6) | 66.8 (−0.6) | 66.0 (−1.0) |
| L1 + R-margin ($N = 32$) | 77.0 (+1.2) | 76.4 (+0.8) | 66.2 (−1.2) | 66.0 (−1.0) |
| L1 + R-margin ($N = 24$) | 76.8 (+1.0) | 74.2 (−1.4) | 65.4 (−2.0) | 67.0 (0.0) |

# F What remains to be measured

In the future we aim to investigate the following:

- Energy and benchmarks for persistence-aware OEA, and benchmarks at larger batches with more items.
- Free generation rather than teacher forcing; other text sources; a second fine-grained model; other GPU generations.
- A kernel that exploits stable experts, with forced changes of expert set at controlled intervals, to price persistence when weights stay in GPU memory.
- Energy shares rather than time shares for the MARL network, by replaying recorded inputs at fixed clocks, with an external meter as a cross-check.
- Energy to reach a target win rate, which is the fair comparison across batch sizes, since larger batches change sample efficiency.
- Narrow-GRU energy; at least 5 seeds with interval estimates; a deliberation cost on the selector [21] or finite scalar quantisation [35] against selector collapse; tasks with distinct phases.